# Depression and Anxiety Prediction Using Deep Language Models and Transfer Learning

Tomasz Rutowski
Ellipsis Health Inc
San Francisco, USA
tomek@ellipsishealth.com

Elizabeth Shriberg
Ellipsis Health Inc
San Francisco, USA
liz@ellipsishealth.com

Amir Harati
Ellipsis Health Inc
San Francisco, USA
amir@ellipsishealth.com

Yang Lu
Ellipsis Health Inc
San Francisco, USA
yang@ellipsishealth.com

Piotr Chlebek
Ellipsis Health Inc
San Francisco, USA
piotr@ellipsishealth.com

Ricardo Oliveira
Ellipsis Health Inc
San Francisco, USA
ricardo@ellipsishealth.com

*Abstract*— Digital screening and monitoring applications can aid providers in the management of behavioral health conditions. We explore deep language models for detecting depression, anxiety, and their co-occurrence from conversational speech collected during 16k user interactions with an application. Labels come from PHQ-8 and GAD-7 results also collected by the application. We find that results for binary classification range from 0.86 to 0.79 AUC, depending on condition and co-occurrence. Best performance is achieved when a user has either both or neither condition, and we show that this result is not attributable to data skew. Finally, we find evidence suggesting that underlying word sequence cues may be more salient for depression than for anxiety.

## I. Introduction

Depression and anxiety are prevalent behavioral health conditions, with significant impact on society [1] and with high rates of co-occurrence [2][3]. These conditions also commonly occur with other health or behavioral health conditions e.g. [4]. Therapies are available for both depression and anxiety [5], but there is a pressing need for screening, so that patients can be identified early [6][7][8]. A growing line of studies has shown that machine-based analysis of spoken language offers promise for the remote detection of behavioral health conditions [9][10]. Spoken language is natural, engaging, and requires only a microphone.

Studies of speech technology with regard to depression prediction in particular has yielded good results for both language models [11][12] and acoustic models [9][10][13][14][15][16]. Advances in machine learning have been supported by common evaluations for depression, including [17][18]. There has been less work on anxiety in the technology community to date.

We are interested in machine learning based approaches for both depression and anxiety when information about their co-occurrence is available. To this end, we introduce a corpus jointly labeled for depression and anxiety, comprising roughly 11,000 unique speakers.

Although the corpus is currently proprietary, it is used because we found no large data public collections that contain both depression and anxiety labels [19] for spoken language. Current data collections are either small or self-reported, with challenges involving the induction of labels based on social media statements [20][21][22][23][24].

We first describe the corpus, and report on the rates of co-occurrence of depression and anxiety as a function of the level of severity for each condition. We then introduce a transfer-learning based deep NLP system. The approach is used to compare performance for the prediction of depression only, the prediction of anxiety only, and the prediction of their joint occurrence. Finally, we define a measure of model variability as a way with which to examine the relative relevance of word information for the prediction of depression versus anxiety.

## II. Method

### A. Data

We use a corpus collected by Ellipsis Health. To facilitate comparative research we have initiated discussions with the Linguistic Data Consortium on a future release of portions of this corpus to the larger speech community [25]. An earlier, smaller version of the data was used to study depression in [26]. Corpus statistics are given in Table I. The data set contains roughly 16,000 sessions from over 11,000 unique speakers. Train and test partitions used for our experiments contain no overlapping speakers. There are more sessions than speakers since some users used the system multiple times; all repeat users are in the training set.

The data is comprised of American English spontaneous speech, with users allowed to talk freely in response to questions within a session. Users range in age from 18 to over 65, with a mean of roughly 30. They interacted with a software application that presented questions on different topics, such as "work" or "home life". Responses average about 175 words. Users responded to 4-6 (mean 4.52) different questions per session, resulting in session durations that averaged just under six minutes each.

During each session, the user completed a PHQ-8 [27] (PHQ-9 after the suicidality question was removed), and a GAD-7. These serve as the depression and anxiety gold standard session labels, similarly to many other research papers [17]. (Since sessions are only a few minutes, it is assumed users do not change state within them.) For both PHQ-8 and GAD-7 scores were mapped for binary classification. Following [28], scores at or above 10 mapped to presence of the condition; scores below 10 were mapped to absence of the condition.

TABLE I
CORPUS STATISTICS, WHERE + / - DENOTE ABSENCE / PRESENCE OF CONDITION BASED ON A THRESHOLD OF 10

|  | *Total* | *Train-* | *Train+* | *Test-* | *Test+* |
|---|---|---|---|---|---|
| **PHQ-8** | | | | | |
| *Responses* | 72,369 | 41,558 | 16,277 | 11,395 | 3,139 |
| *Sessions* | 15,950 | 9,266 | 3,606 | 2,425 | 653 |
| **GAD-7** | | | | | |
| *Responses* | 72,369 | 42,662 | 15,173 | 11,539 | 2,995 |
| *Sessions* | 15,950 | 9,538 | 3,334 | 2,460 | 618 |

*B. NLP System*

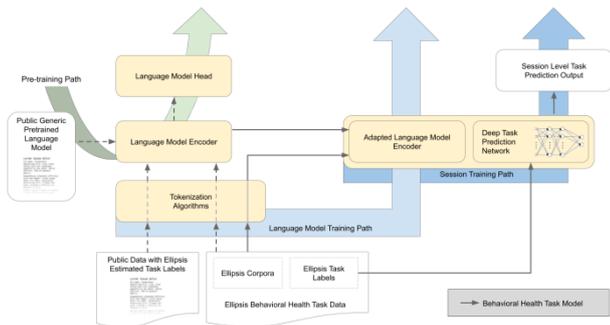

Fig. 1 NLP Process Flow

Learning word representations has been well known for many years. One of the most relevant approaches was based on building the representation in a vector space that enabled to group similar words together. Initially, in our research we used word embeddings [29][30] and SVM classification algorithm. Another direction was to use a transfer learning approach [31] where the entire language model was used for classification purposes.

The process flow in Fig. 1 starts from building a generic language model and adapting it for our task. Currently, the two most common topologies encompass recurrent networks and transformers. Due to a small memory and computation footprint of the recurrent networks topology, still maintaining very high accuracy, it has been used very extensively in our research. Various methods can be used for tokenization purposes [33][34]. We use the spaCy tokenization library; each word is represented by a unique ID. Our word dictionary contains over 20,000 individual tokens. Due to the computing limitations and the time needed to train language models from scratch the core topology as well as tokenization technique is usually provided by third party entities (in order to train some of the language models within a few days constraint at least dozens of GPUs are needed). Most of the available NLP generic language models are trained on publicly available data e.g. web crawls, books, Wikipedia [32]. Our process encompasses a middle step in which a generic language model is further pre-trained on a public data that we specifically collected in order to put an emphasis on emotion and mental health domain language structure. We find that this approach stabilizes final depression predictions. This study is primarily based on LSTM topology inspired by the AWD-LSTM architecture [35] of the core language model and ULMFiT [36] work for model fine tuning. The approach in [35] is based on the following contributions: DropConnect for hidden to hidden layers which differs from the drop out approach by deactivating certain weights rather than activation mechanisms. Another beneficial technique, back-propagation through time; dynamically changes the sequence length for the forward and backward passes. Embedding dropout: where occurrences of certain words are gone during the training stage. Furthermore, the approach in [36] primarily leverages the likes of transfer learning: the language model is trained on a large data corpus none related to the downstream task. We then start an adaptation step where the proprietary data without labels is used to further train Adapted Language Model encoder. As a next step, additional layers are added to it for classification or regression purposes (Prediction Model). The key mechanisms within this implementation encompass discriminative fine tuning, where the different layers of the network use a different learning rate. This rate is the slanted triangular learning rate, which is where the rate of change is dependent on the stage of the training process. The remaining main features cover gradual unfreezing, back-propagation through time to handle longer language dependencies and concatenated pooling of multiple time steps from the recurrent network.

### III. LABEL DISTRIBUTION AND CO-OCCURRENCE

We first examined the distribution of depression, anxiety, and their co-occurrence in our data. As noted in Section II, labels for each behavioral health condition come from self-report surveys and are quantized using a threshold of 10, after [17], into binary classes. We use dep+ or anx+ where the score for the noted condition is 10 or greater. To denote the co-occurrence patterns, we first list the binary value for depression and then for anxiety. For example we use "+,+" to denote that PHQ-8 and GAD-7 are each at or above 10, and "-,-" to denote that both are below 10. Table II provides information on the joint distribution of labels. Initial analysis encompasses overlapping labels and the correlation between +,+ vs -,-. The marginal rates of depression and anxiety are both on the order of 25% positive cases. Overall, about 18.5% of the sessions are positive for both labels. In total, roughly 15% (~2,000) of the sessions are single-condition, meaning either the user was positively diagnosed as depressed or anxious, but not both depressed and anxious. For the test subset the proportions are roughly the same; joint-positive plus joint-negative samples represent 88%, vs. 85% in the full data set.

TABLE II
LABEL DISTRIBUTION BY BINARY CLASS. COUNTS ARE THE NUMBER OF SESSIONS IN THE FULL CORPUS AND TEST SET.

|  | *anx+* | *anx-* |
|---|---|---|
| *dep+ Full* | **2,964 (18.5%)** | **1,295 (8.1%)** |
| *dep+ Test* | 455 (14.7%) | 198 (6.4%) |
| *dep- Full* | 988 (6.1%) | 10,703 (67.1%) |
| *dep- Test* | 163 (5.2%) | 2,262 (73.4%) |

As shown, most of the speakers had either neither condition, or both conditions. While our data set is relatively large compared with past speech corpora for these conditions, the amount of test data for the speaker having only one condition, is below 200 test sessions. For this reason we do not use 4-way classification in the present study; such work awaits future additional test data.

Fig. 2 represents a normalized histogram (in line plot format) for each survey (gold standard) value. The PHQ and GAD histograms are nearly identical, except for label values at 0, where there is a 5% discrepancy (800 sessions). Fig. 2 is for the entire data set (train and test). We observe a similar pattern in the test set only.

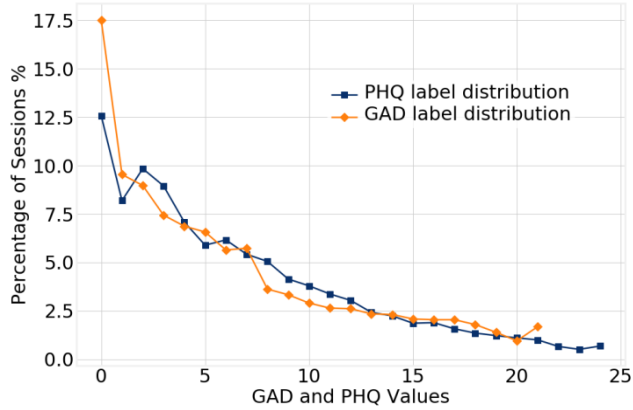

Fig. 2 Distribution of label values in corpus

The overall correlation between GAD and PHQ in our train and test set is high, at 0.80 [37]. Another way of looking at the label distribution is the matrix of each label score relative to the other, as shown in Fig. 3. Note the difference in score ranges. Each survey instrument question offers 4 possible scores (0, 1, 2, and 3). GAD-7 scores thus range from 0 to 20; PHQ-8 scores range from 0 to 24. Within each scale, higher values indicate higher condition severity. Fig. 3 represents the counts of sessions for a given combination of two specific label values. As shown, the majority of sessions occur near the diagonal, consistent with high correlation of two mental health conditions [38][2]. Interestingly, there is a greater variation of PHQ labels for each GAD label rather than vice versa. That is, there is more variability in rows than in columns. This deserves further study but may reflect an asymmetry in how depression and anxiety behave as preconditions; anxiety tends to be a precondition for depression more often than vice versa [2].

Fig. 3 Co-occurrence of PHQ-8 and GAD-7 by severity level, in the corpus.

## IV. DEEP NLP PERFORMANCE

### A. Model accuracy by condition

Before looking at the binary classification results, we analyzed model accuracy as a function of raw PHQ and GAD scores prior to their mapping for binary classification. As shown Fig. 4, there is a large variation in accuracy depending on score value distribution. Values near the ends of the score range have accuracies near 0.90, whereas those near the threshold of 10, which are expected to be more confusable, are closer to 0.50. The noise at higher scores reflects an increase in data sparsity.

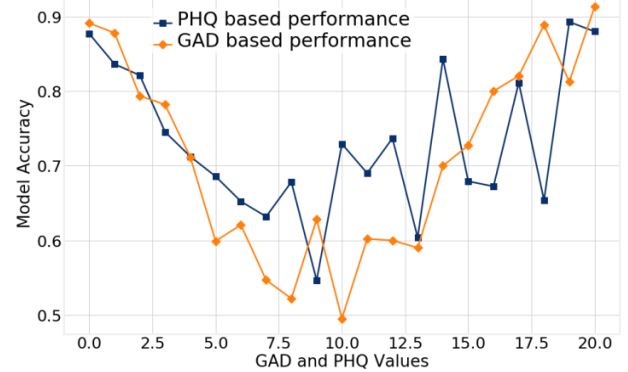

Fig. 4 Model accuracy by condition severity

The model has been trained for classification using binary classes. For both GAD and PHQ values at 10 or higher are mapped to positive for the condition. Our model achieves an AUC score for depression of 0.828. The result for anxiety is slightly lower, at 0.792. One possible reason for the difference is simply less attention so far in our work to model tuning for anxiety than depression. A second, independent possibility is discussed in *Section C*. As described in that section, there is some evidence for a fundamental difference in language sequence cues for prediction of anxiety, as compared with similar types of cues for depression. *Section C* discusses this possibility, as well as a countering effect of the condition priors, that may help explain these performance differences.

TABLE III
MODEL PERFORMANCE BY CONDITION

| | AUC | Specificity (at EER) | Sensitivity (at EER) |
|---|---|---|---|
| *PHQ* | 0.828 | 0.755 | 0.755 |
| *GAD* | 0.792 | 0.722 | 0.721 |

Given the large literature on depression detection in common corpora noted in Section I, we also seek to provide comparison information on our performance. Although we did not have access to the common data, we provide the following results to allow informal, indirect comparison. Leading, recent results on the AVEC common task are described in [17]. The AVEC data and Ellipsis data differ, but both use PHQ-8 scores as labels. We report metrics found in [17], using regression. The test results in [17] for RMSE, an error metric, are 5.51 versus 4.21 for our study. We believe we are thus in the range of competitive performance.

## B. Co-occurrence effects

We discovered that model performance is best when the speaker has either both conditions present (+,+), or both conditions absent (-,-). AUC for discriminating joint-presence and joint-absence sessions increases to 0.861 and 0.845, for PHQ and GAD respectively. We then asked whether this effect could be simply due to a change in session priors. The priors for joint- presence and joint-absence data change from around 0.20 to 0.16. After rebalancing the data (changing priors from 0.167 to 0.211), we found that this is not the case. The improved result remained—and even increased, to 0.863 and 0.849 respectively. This suggests that class discrimination is better for joint modeling of depression and anxiety than for individual modeling of either condition.

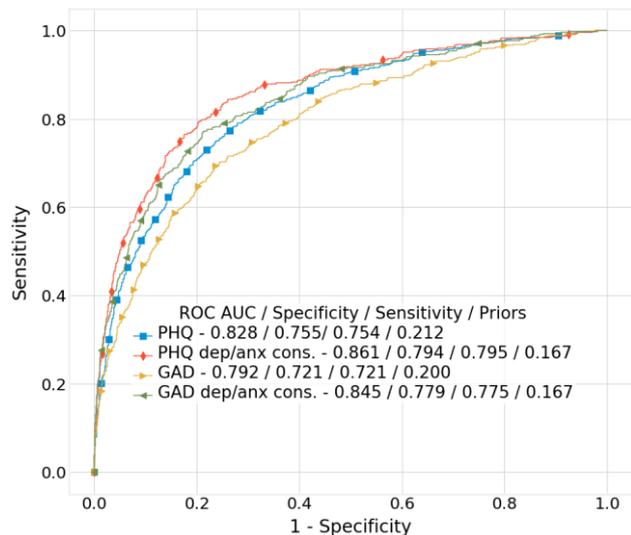

Fig. 5 AUC and EER-based sensitivity and specificity results. The legend is ordered by ascending AUC performance within each condition

## C. Within-session model variability by class

Given the nature of the models used (see Section 2.2), we assume that certain word sequences and their dependencies are used as cues for the models, to separate positive from negative cases for each condition. To investigate, we gated the word sequence in a forward direction, to estimate the amount of prediction information available at a given time during a test session. For example, in the session of 800 words, the same number of cumulative gated samples was generated by adding one word at a time starting from the first token; in this case, 800 predictions were returned. For our total of 3078 test sessions, roughly 2.4M predictions were generated. Based on these predictions, we calculated a value we call Within-Session Model Variability. This process was performed separately for each condition. Note that the test set is exactly the same word sequence data (from the same speech), for both models.

Table IV provides results for this measure of variability for the depression model. As shown, variability within a session is highest for +,+ (both conditions), lowest for -,- (neither condition), and in between for the mixed cases. This suggests that a model tuned for maximum AUC on binary depression classification, is using word sequence cues associated with higher variability on this measure within a session.

TABLE IV
WITHIN-SESSION (W-S) MODEL VARIABILITY BY CROSSED CONDITION, FOR THE PHQ (DEPRESSION) MODEL

| W-S Model Variability | dep+ | dep- |
|---|---|---|
| **anx+** | 0.090 | 0.088 |
| **anx-** | 0.084 | 0.077 |

The model trained for anxiety shows a similar pattern; see Table V. Here, however, (1) overall variability is lower than that for depression, and (2) variability for the (-,-) case is much lower than expected given the other three values. Because the same test data is used for both tables, and the NLP model methods are the same, this suggests that the word sequence cues for anxiety may be weaker or less prevalent, than those for depression.

TABLE V
WITHIN-SESSION (W-S) MODEL VARIABILITY BY CROSSED CONDITION, FOR THE GAD (ANXIETY) MODEL

| W-S Model Variability | dep+ | dep- |
|---|---|---|
| **anx+** | 0.077 | 0.065 |
| **anx-** | 0.061 | 0.048 |

A question, then, is why anxiety performance in AUC is only slightly below that for depression, as shown in Fig. 5 and earlier. The answer may lie in the difference in severity distributions. Fig. 2, Fig. 3 and Fig. 4 together, indicate that in our data, there is a preponderance of labels with a score of 0 on anxiety. Because this value is far from the binary classification threshold, the overall AUC performance for anxiety is boosted, compared with that for depression.

## V. SUMMARY AND CONCLUSIONS

We have described a large corpus of spontaneous speech, labeled jointly for depression and anxiety. The data show a high level of co-occurrence of the two conditions, as well as a positive data skew for severity within each condition. We examine performance of deep language models that make use of transfer learning for model pre-training. We achieve a classification performance of 0.83 AUC for depression, and 0.79 for anxiety; these results increase to AUC=0.86 and 0.85 respectively, for sessions with either both conditions or neither condition (+,+ and -,-) which was the primary objective of the paper. After adjusting for data skew, these results persist, suggesting that discrimination is better for joint modeling. In addition, we examine within-session model variability. Because our data are matched over conditions, results suggest that word cues for depression may be fundamentally stronger than those for anxiety. Taken together these results reveal the importance of considering condition co-occurrence for speech- based behavioral health screening and monitoring applications.